\documentclass{article}

\usepackage{arxiv}

\usepackage[utf8]{inputenc} % allow utf-8 input
\usepackage[T1]{fontenc}    % use 8-bit T1 fonts
\usepackage[pagebackref,breaklinks,colorlinks]{hyperref}
\usepackage{url}            % simple URL typesetting
\usepackage{booktabs}       % professional-quality tables
\usepackage{amsfonts}       % blackboard math symbols
\usepackage{nicefrac}       % compact symbols for 1/2, etc.
\usepackage{microtype}      % microtypography
\usepackage{lipsum}

\usepackage{amssymb}
\usepackage{amsmath}
\usepackage[normalem]{ulem}
\usepackage{multirow}
\usepackage{array}
\usepackage{graphicx}
\usepackage{bm}
\usepackage{caption,subcaption}        % for subfigures
\usepackage{pifont}
\usepackage{xcolor}                    % use \textcolor{red}{text}

\title{PLSP (Pre-hoc Liminal Space Profiling): OOD Prediction over Detection -- An Anticipatory Approach for Machine Learning Model Reliability}

\author{
    Vipul Bansal \\
    Indian Institute of Technology Roorkee \\
    Roorkee, India \\
    \And
    Himanshu Buckchash\thanks{
    \copyright{} 2026 EURASIP. Accepted for publication in the Proceedings of the 34th European Signal Processing Conference (EUSIPCO 2026), published by EURASIP.
    This work was supported in part by MeitY, Govt.\ of India, and IMC University of Applied Sciences Krems, Austria. Vipul~Bansal’s contributions were made at IIT Roorkee, India. Correspondence can be made to: himanshu.buckchash@imc.ac.at.} \\
    IMC University of Applied Sciences \\
    Krems, Austria \\
    \And
    Balasubramanian Raman \\
    Indian Institute of Technology Roorkee \\
    Roorkee, India \\
    \And
    Deepak Dhungana \\
    IMC University of Applied Sciences \\
    Krems, Austria \\
}

\begin{document}
\maketitle

\begin{abstract}
Out-of-Distribution (OOD) data poses a significant threat to machine learning models, often leading to model failure during deployment. All existing OOD detection methods are post-hoc, relying on evaluation metrics such as accuracy and AUC-ROC during inference to indirectly assess the model's response to OOD data by measuring deviations.
In contrast to existing approaches, the proposed work shifts the paradigm from \textit{OOD detection to OOD prediction} by proposing a pre-hoc anticipatory framework called PLSP for OOD prediction. We make several key contributions: (a) a dataset-independent metric called the CREDibility Score (\textit{CREDS}) is proposed for OOD prediction; (b) credibility curves are introduced to study the maximum credibility a model can attain; and (c) credibility heat maps (and volume under surface) are introduced to characterize pre-hoc model behavior across different datasets.
This work provides a novel perspective on signal processing under distributional shifts. Experiments across multiple datasets demonstrate that the proposed metric serves as a valuable measure for improving the robustness of machine learning models toward OOD prediction.
\end{abstract}

% keywords can be removed
\keywords{out-of-distribution detection \and OOD prediction \and model robustness \and distribution shift \and machine learning reliability \and credibility metrics}

\section{Introduction}
Many real world critical applications of machine learning such as medical image analysis or anomaly detection,
%, open-set recognition, novelty detection, outlier detection,
share the common challenge of (inference-time) distribution shift in various degrees \cite{yang2021generalized}. This problem arises due to the presence of OOD data during inference due to regional or environmental changes \cite{roy2021does,gulrajani2020search}. This leads to reduced performance, increased safety risk, reduced customer satisfaction, and revenue and trust loss to the service provider \cite{fort2021exploring}.

Main reason behind this challenge is the inability of machine learning methods to exhaustively model the underlying data distribution \cite{krizhevsky2012imagenet,he2015delving,drummond2006open}. Supervised models are often trained by minimizing the generalization error with the assumption that the train and test samples are drawn from an i.i.d. distribution \cite{liu1995unbiased}. This assumption however, does not hold on the real world data due to the presence of out-of-distribution (OOD) samples \cite{fort2021exploring,drummond2006open,winkens2020contrastive,milbich2021characterizing,thanh2020toward}.

In the existing literature, the generalization performance is measured either using metrics like average or worst-case accuracy over multiple OOD datasets \cite{frogner2021incorporating,shen2020stable,liu2021heterogeneous,zhou2021domain}, or through a distance metric over a single OOD dataset \cite{fort2021exploring,winkens2020contrastive,thanh2020toward,ye2021out,milbich2021characterizing,ye2021ood,neyshabur2017exploring}. However, these approaches, do not directly perceive the importance of features during OOD data evaluation. Due to this, they are not able to assess the impact of changes on the model for any given OOD data sample \cite{arjovsky2020invariant,krueger2021out,xu2020adversarial,yan2020improve}. These approaches also lack the ability to measure generalization of the model for any potential change in OOD input. However, the proposed \textbf{CREDS} metric perceives both \textbf{feature level understanding} of a model and \textbf{distributional changes}. It enables the assessment of the maximum possible credibility in case of variation in the form of a given OOD dataset and also in absence of an OOD dataset.

In this work, \textit{\textbf{Credibility} is defined as the measure of generalizability of a model}. High credibility implies higher generalization ability of the model, similarly, low credibility implies low generalization capacity of the model.

\begin{figure}[t]
    \centering
    \includegraphics[width=.85\columnwidth]{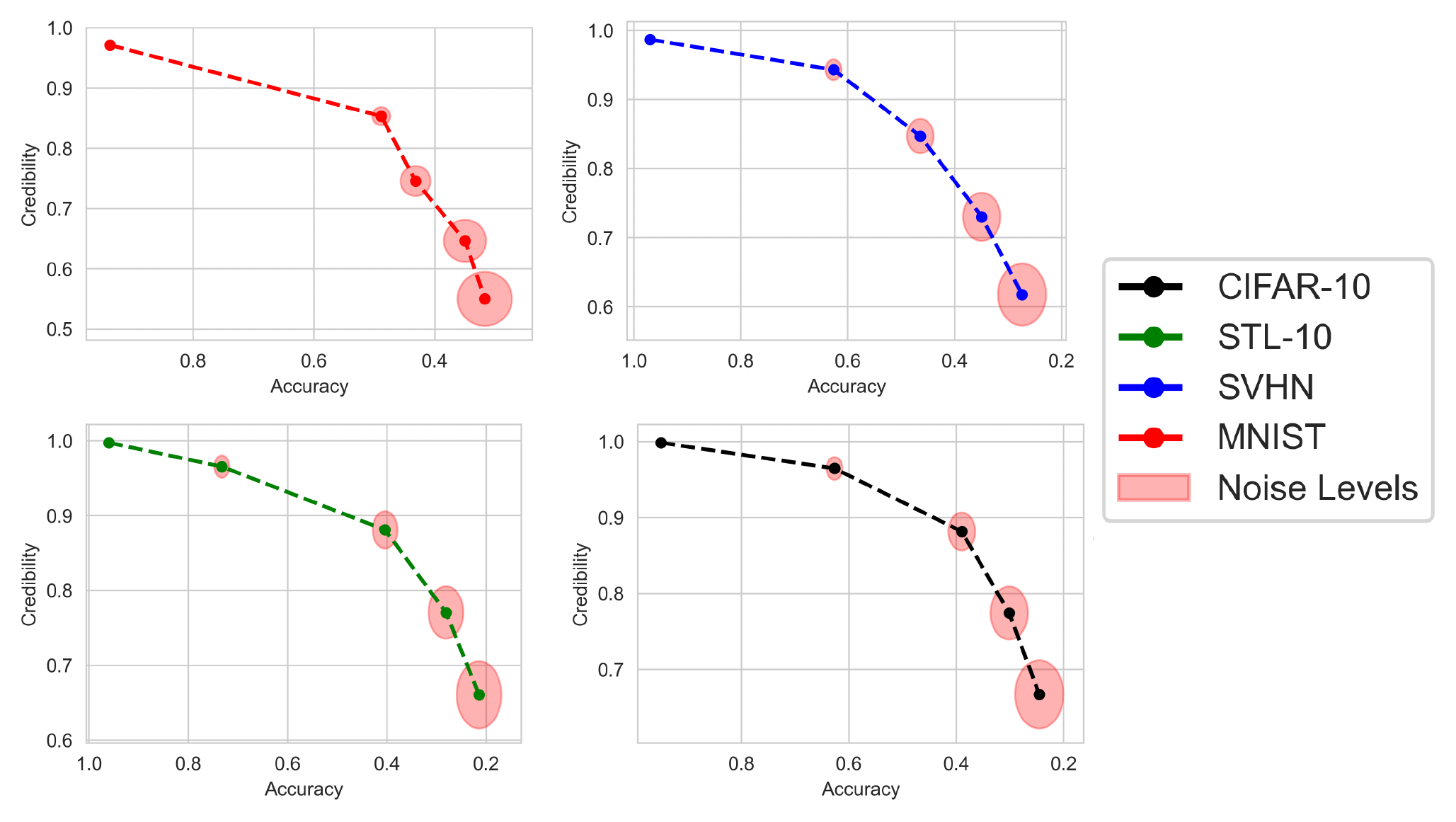}
    \caption{Direct correlation between CREDS and accuracy across increasing noise levels (0-50\%) for different datasets, demonstrating CREDS as an effective competitor to post-hoc metrics.}
    \label{fig:main}
\end{figure}

Our findings reveal that the proposed pre-hoc CREDS metric provides estimates comparable to the post-hoc accuracy metric, as evidenced by the monotonous decline in both measures with increasing noise across all datasets (Fig.~\ref{fig:main}). This correlation demonstrates that CREDS serves as a valid measure of model trustworthiness and generalizability. Critically, unlike existing post-hoc metrics, CREDS can be evaluated \textbf{in the absence of any OOD dataset(s)}, offering the distinct advantage that model behavior under distributional shifts can be studied and adjusted during the development environment before deployment, thereby preventing catastrophic failures in production settings.

\textbf{CREDS operates} by first measuring how much the statistical distribution of each input feature in new data deviates from its training-data counterpart, for each specific class. This deviation is converted into a foundational credibility score for the feature and class, subsequently weighted by the model's reliance on that feature. Finally, these model-aware feature credibilities are aggregated across all features and then averaged over all classes to produce the overall CREDS score.

The main contributions of this work are:
\begin{itemize}
	\item To the best of our knowledge, we are the first to propose a direct measure (CREDS) of a machine learning model's generalization ability and show its application. It is an approximate formulation, under assumptions, since a closed form solution is not possilbe.

	\item We also propose a method to assess possible changes in the credibility of a model upon distributional changes due to OOD data during inference. The proposed AUC metric for change in credibility, helps us understand generalization of a model for a particular OOD dataset. We also propose a new Volume Under Surface (VUS) metric to measure the extent of model generalization for changes in input distribution, given we don't have any OOD data to evaluate our model.

	\item We conduct experiments on multiple datasets to show the usage of credibility scores and associated methods to evaluate models. We also study the impact of noise on credibility and generalization of a model.
\end{itemize}

\section{Credibility Based OOD Evaluation}
In Section~\ref{sec: cred}, we mathematically introduce CREDS that helps us to evaluate how the model sees an input distribution. It is an approximate formulation (under stated assumptions) which is empirically validated to work. In Section~\ref{sec: curves}, we then introduce credibility curves to understand the change in credibility score with changing mean and deviation. From these curves, we find the maximum possible credibility score, a model can achieve, given changes in an input data stream. Lastly, In Section~\ref{sec: vul}, we evaluate the overall vulnerability of the model for OOD data.

\subsection{Credibility Score (CREDS)}
\label{sec: cred}

The way a model is affected by OOD data depends mainly on two factors. Firstly, \textit{how does the distribution of OOD data vary w.r.t. the training data?} Secondly, \textit{how the changes in feature distribution are perceived by the model?} To understand this, we mathematically define credibility (CREDS), ${C}_{k,j}$ for the $k^{th}$ feature, given a class label $j$ as: 
\begin{equation}
    \label{eq:cred}
	\begin{split}
		C_{k,j} & =e^{-D_{KL}(p(X^v_k|Y=j)||q(X^t_k|Y=j))}\\
		& = e^{-\mathrm{E}_{p(X^v_k|Y=j)}\Big[\log\big(\frac{p(X^t_k|Y=j)}{q(X^v_k|Y=j))}\big)\Big]}\\
		% &= e^{\big\int_{-\infty}^{\infty} {p(X^v_k/Y=j)\Big[\log\big(\frac{q(X^v_k/Y=j))}{p(X^t_k/Y=j)}\big)\Big] d} }\\
	\end{split}
\end{equation}
Where $X^t_k$ and $X^v_k$ are $k^{th}$ feature for training and OOD dataset respectively, and $q$ and $p$ are the respective probability distributions. Here, $D_{KL}$ represents the KL divergence between the two distributions \cite{Kullback59,zhang2021properties}. If we consider $p$ and $q$ as multivariate normal distributions with mean $\mu_{v,k}$ and  $\mu_{t,k}$ and covariance matrix $\Sigma_{v,k}$ and $\Sigma_{t,k}$ respectively with dimensions $\lambda$, then we can compute: 
\begin{equation}
	\begin{split}
		D_{KL}(p||q) & = \frac{1}{2}\Big\{tr\{\Sigma_{t,k}^{-1}\Sigma_{v,k}\}
		\\ & + (\mu_{t,k} - \mu_{v,k})^T\Sigma_{t,k}^{-1} (\mu_{t,k} - \mu_{v,k} )\\ 
		& - \lambda + \ln{\frac{|\Sigma_{t,k}|}{|\Sigma_{v,k}|}}\Big\}
	\end{split}
\end{equation}
For simplicity of calculation (for tractability and feasibility) we can estimate deviation $\sigma_{v,k}$ for the dataset instead of covariance $\Sigma_{v,k}$. Assuming a normal distribution, we can rewrite $D_{KL}$ as:

\begin{equation}
	\label{eq:KL}
	D_{KL}=\Big[\ln\big(\frac{\sigma_{t,k}}{\sigma_{v,k}}\big)+\frac{\sigma_{v,k}^2+(\mu_{v,k}-\mu_{t,k})^2}{2\sigma_{t,k}^2}-\frac{1}{2}\Big]
\end{equation}

Even if the true distributions aren't perfectly Gaussian, changes in their central tendency (mean) and dispersion (variance) are often the most significant and impactful types of shifts for many ML models. The Gaussian $D_{KL}$ formula is sensitive to exactly these. Moreover, by calculating $C_{k,j}$ for each feature and class, CREDS can pinpoint where the distributional shift is occurring at a granular level.

The credibility $C_{k,j}$ just accounts for distribution difference in between OOD data and training data. To incorporate how model perceives the distributional changes, we take feature importance $f_k$ into account to calculate class credibility  $C_j$ as: 
	\( C_j=\sum_k f_k C_{k,j} \).
To estimate a feature's overall effects on the results, we can also estimate $C_k$ for a given feature as: 
	\( C_k=\sum_j \frac{n_j}{\sum_j n_j} C_{k,j} \).
Where $n_j$ are the number of samples in class $j$ of training dataset. 
We can further compute the overall credibility score $C$ of the model as: 
\begin{equation}
	C=\sum_j\sum_k \frac{n_j}{\sum_j n_j} f_k C_{k,j}
\end{equation}
Overall, credibility $C$ is a collective measure of:  ``\textit{How model observes input distribution weighted over all the classes and features}".  

The credibility score (CREDS) is \textbf{not a performance score} like Accuracy or AUC-ROC. Instead it tells us about feasibility to utilize our model to predict on given test data. The metric helps to understand tendency to generalize under unknown distributions. This includes both data and model's perspective on perceiving the data.

\subsection{Credibility Curves}
\label{sec: curves}
To evaluate the changes observed in credibility score with mean and deviation of input data stream we propose utilization of credibility curves. These curves show changes in $C_j$ for a class $j$ by fixing either $\mu_{v,k}$ or $\sigma_{v,k}$ for all $k$ features.

To assess generalization of credibility for a class $j$ with given set of $\sigma_{v,k}$, we vary the value of $\mu_{v,k}$ for each feature in a range $-\epsilon$ to $+\epsilon$ given $\epsilon>0$. We calculate Area Under the $C_j$ vs $\mu_v$ curve for a given class $k$ to estimate the potential of network as:
\begin{equation}
	AUC{\mu_v}^\epsilon_j = \int_{-\epsilon}^{\epsilon} \sum_k f_k C_{k,j}(\mu_{t,k},\mu_{v},\sigma_{t,k},\sigma_{v,k}) d\mu_v
\end{equation}

\begin{equation}
	AUC{\mu_v}^\epsilon =\sum_j \frac{n_j}{\sum_j n_j}  AUC{\mu_v}^\epsilon_j
\end{equation}

where $C_{k,j}(.)$ function gives estimate of $C_{k,j}$ for a set of parameters. Similarly, we calculate Area Under the $C_j$ vs $\sigma_v$ curve for a given class $k$ to estimate the potential of network as:
\begin{equation}
	AUC{\sigma_v}^\epsilon_j = \int_{0}^{\epsilon} \sum_k f_k C_{k,j}(\mu_{t,k},\mu_{v,k},\sigma_{t,k},\sigma_{v}) d\sigma_v
\end{equation}
\begin{equation}
	AUC{\sigma_v}^\epsilon =\sum_j \frac{n_j}{\sum_j n_j}  AUC{\sigma_v}^\epsilon_j
\end{equation}

Higher the value of $AUC{\mu_v}^\epsilon_j$ or $AUC{\sigma_v}^\epsilon_j$, more credibility one model can sustain for each class $j$ on potential changes in $\mu_v$ or $\sigma_v$.  
We also estimate the maximum credibility $C^\mu_{j,max}$ and $C^\sigma_{j,max}$ for variation of $\mu_v$ and $\sigma_v$ respectively. For that we estimate $\partial C_{k,j} / \partial \mu_v$ and $\partial C_{k,j} / \partial \sigma_v$ as: 
\begin{equation}
	\frac{ \partial C_{j} }{ \partial \mu_v } =\sum_k f_k \frac{ \partial C_{k,j} }{ \partial \mu_v }= -\sum_k f_k \frac{(\mu_v-\mu_{t,k})}{\sigma_{t,k}^2} C_{k,j}
\end{equation}

\begin{equation}
	\frac{ \partial C_{j} }{ \partial \sigma_v } =\sum_k f_k \frac{ \partial C_{k,j} }{ \partial \sigma_v }= \sum_k f_k  \Big[\frac{1}{\sigma_{v}} - \frac{\sigma_{v}}{\sigma_{t,k}^2} \Big] C_{k,j}
\end{equation}
For a maxima we estimate $C_j$ such that $\partial C_j / \partial \mu_v =0$ and $\partial^2C_j / \partial^2\mu_v < 0$ and $\partial C_j / \partial \sigma_v= 0$ and $\partial^2C_j / \partial^2\sigma_v < 0$. For the case, given $\mu_{t,k}= \mu_t \forall k$, one possible solution for same is $\mu_v=\mu_t$. Other possible way to find a maxima, is to generate samples for range of values for $\mu_v$ and $\sigma_v$ and estimate $C^\mu_{j,max}$ and $C^\sigma_{j,max}$.
These curves help to see how change in current distribution affects the credibility score of the potential data.

\subsection{Model Vulnerability Analysis}
\label{sec: vul}

If we don't have an available OOD dataset, we would still like to see how our model behave to potential OOD distributions. To do so we plot a heatmap of $C_j$ with variation in both $\mu_{v,k}$ and $\sigma_{v,k}$ such that  $\mu_{v,k}=\mu_v \forall k$ and $\sigma_{v,k}=\sigma_v \forall k$. These plots help us to understand variation of credibility for a class with distribution of data.

Similar to $AUC_{\mu_v j}^\epsilon$ and $AUC_{\sigma_v j}^\epsilon$ we can compute the Volume Under the Surface (VUS) of these plots between $C_j$, $\mu_v$ and $\sigma_v$ to see how well the model performs for a class when there is a change in distribution. Mathematically we can define it as: 
\begin{equation}
	VUS_j^{\alpha,\beta} = \int_0^{\alpha} \int_{-\beta}^{\beta} \sum_k f_k C_{k,j}(\mu_{t,k},\mu_{v},\sigma_{t,k},\sigma_{v})d\mu_v d\sigma_v 
\end{equation}
\begin{equation}
	VUS^{\alpha,\beta} =\sum_j \frac{n_j}{\sum_j n_j}  VUS_j^{\alpha,\beta} 
\end{equation}
Where $VUS^{\alpha,\beta}$ gives an overall metrics for a model. Higher the value of $VUS_j^{\alpha,\beta}$, more chances for a class to achieve better credibility for potential OOD data. If all the features of the dataset have a $\mu_{t,k}=0$ and $\sigma_{t,k}=1$, then there is atleast one point for which $C_j$ is exactly equal to 1.

\section{Experiments}\label{sec: Experiment}
This section demonstrates how these methods can be utilized for evaluating various cases. 
We evaluate on five benchmark datasets spanning handwritten digits (MNIST \cite{MNIST}, SVHN \cite{SVHN}) and natural images (CIFAR-10 \cite{cifar}, CINIC-10 \cite{CINIC}, STL10 \cite{STL}), with training samples ranging from 500 to 90K. All images are preprocessed into feature vectors of size 49 (digit datasets) or 64 (natural image datasets) through dimensionality reduction. This diverse collection enables comprehensive evaluation across different data distributions and complexities. Additionally, \textbf{Noisy Datasets} are created to see how increasing noise impacts the model credibility. Four noise levels are used: $10\%$, $20\%$ ,$30\%$ ,and $40\%$.

\subsubsection{\textbf{Training Details}} We run experiments by pairing MNIST with SVHN and CIFAR-10 with CINIC-10 and STL-10. The dataset MNIST and SVHN form a pair of dataset which are out of distribution for each other. CIFAR-10, CINIC-10 and STL-10 forms a group datasets which is partially different from each other. We train a random forest \cite{RF} model on features and utilize the feature importance received from the model  to calculate credibility scores (CREDS), $C$. We further estimate credibility curves by changing $\mu_v$ and $\sigma_v$ and estimate $C^\mu_{max}$ and  $C^\sigma_{max}$ for the dataset and also estimate $AUC^5_{\mu_v}$ and  $AUC^5_{\sigma_v}$. For these datasets we also estimate $VUS^{5,5}$.

\begin{table}[t]
    \caption{Overall CREDS results (with mean and deviation) for MNIST and SVHN. Credibility is estimated for all the permutations of SVHN and MNIST dataset. Additionally, noise was varied in each dataset from 0\% to 50 \% to note the change in CREDS.}
    \centering
    % \resizebox{.7\columnwidth}{!}{%
        \begin{tabular}{|l|l|l|}
            \hline
            Train Dataset & Test Dataset & Credibility \\ \hline
            \multirow{6}{*}{MNIST} & SVHN & 0.458 $\pm$   0.085 \\
            & MNIST & 0.971 $\pm$   0.021 \\
            & MNIST(10\% Noise) & 0.853 $\pm$   0.059 \\
            & MNIST(20\% Noise) & 0.745$\pm$  0.072 \\
            & MNIST(30\% Noise) & 0.646$\pm$  0.077 \\
            & MNIST(40\% Noise) & 0.550$\pm$  0.072 \\ \cline{2-3} 
            \multirow{6}{*}{SVHN} & MNIST & 0.444$\pm$  0.083 \\
            & SVHN & 0.986$\pm$  0.003 \\
            & SVHN(10\% Noise) & 0.943$\pm$  0.007 \\
            & SVHN(20\% Noise) & 0.847$\pm$  0.012 \\
            & SVHN(30\% Noise) & 0.730$\pm$  0.015 \\
            & SVHN(40\% Noise) & 0.617$\pm$  0.017 \\ \hline
        \end{tabular}
    % }
    \label{tab:MNIST-SVHN-C}
\end{table}

\subsubsection{\textbf{Credibility on MNIST \& SVHN}}
We train our model on MNIST and test on SVHN and vice versa and observe the credibility score as given in Table~\ref{tab:MNIST-SVHN-C}. Intra-dataset training and testing on MNIST \& SVHN give high CREDS values of $0.971$ \& $0.986$ respectively. Higher CREDS on intra-dataset evaluation, show that the test data acts like an In-Distribution for the model. This is because $\mu_{v,k} \rightarrow \mu_{t,k}$ and $\sigma_{v,k} \rightarrow \sigma_{t,k}$. This brings $D_{KL}$(refer (Equation~\ref{eq:KL})) tend to $0$, in turn bringing $C_{k,j}$(refer (Equation~\ref{eq:cred})) close to $1$.  
When training on MNIST and testing on SVHN, CREDS is $0.4581$, indicating significant dataset differences. The reverse configuration yields $0.4446$, slightly lower due to differences in prior distributions $p(X^t_{k}/Y=j)$, feature importances $f_k$, and class distributions $n_{j}/\sum_j n_j$. Note that $C_{k,j}$, being exponential in $D_{KL}$, captures larger variations at higher $D_{KL}$ values despite small changes in credibility.

\begin{figure}[t]
    \centering
    \includegraphics[width=\columnwidth]{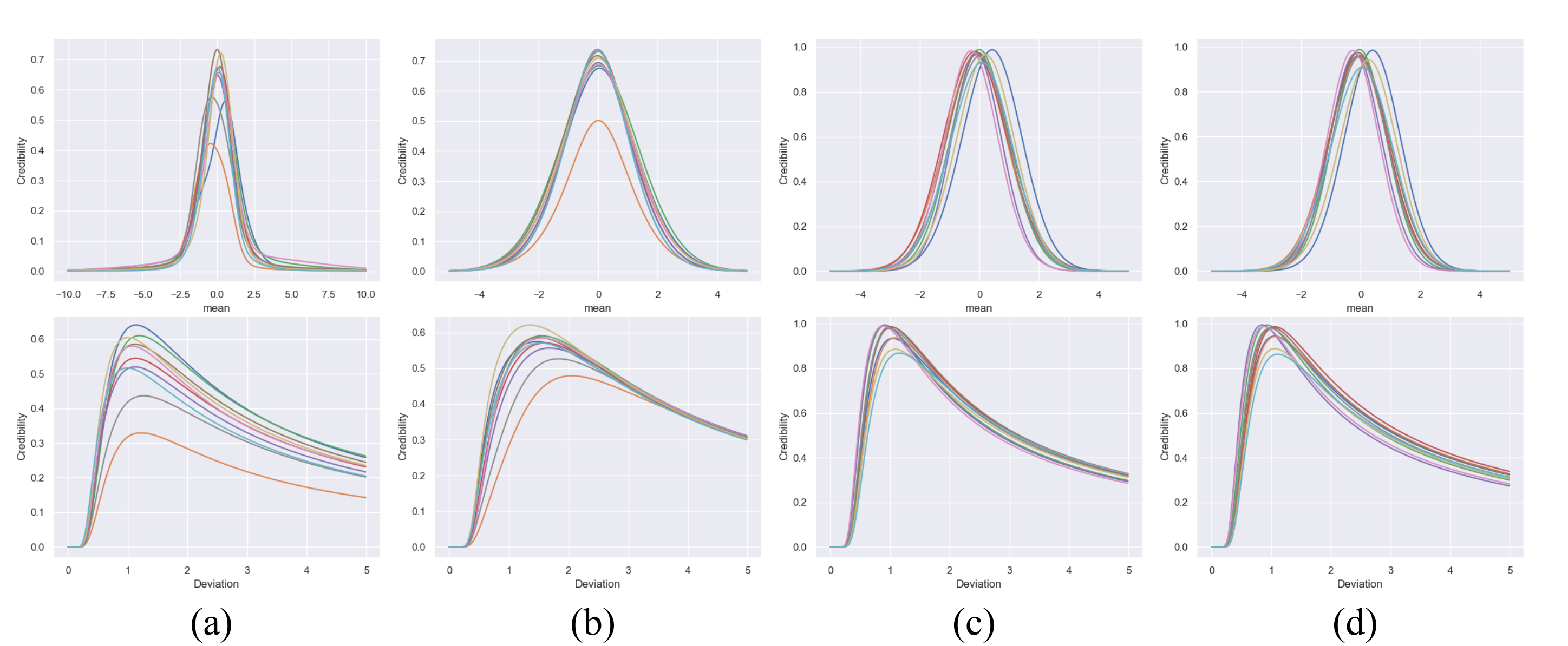}
    \caption{Credibility Curves for variation in $\mu_v$ and $\sigma_v$ for (a) model trained on MNIST and tested on SVHN (b) model trained on SVHN and tested on MNIST (c) model trained on CINIC-10 and tested on CIFAR-10 (d) model trained on CIFAR-10 and tested on CINIC-10.}
    \label{fig:curves}
\end{figure}

\begin{table}[t]
    \caption{Credibility results (with mean and std.) for experiments on CIFAR-10, STL-10, and CINIC-10. Credibility is estimated for all the permutations of all datasets. Noise was varied in each dataset from 0\% to 50 \% to note the change in CREDS.}
    \centering
    % \resizebox{.7\columnwidth}{!}{%
    \begin{tabular}{|l|l|l|}
        \hline
        Train Dataset & Test Dataset & Credibility \\ \hline
        \multirow{7}{*}{CIFAR 10} & CINIC-10 & 0.940$\pm$  0.059 \\
        & STL-10 & 0.892$\pm$  0.080 \\
        & CIFAR-10 & 0.998$\pm$  0.000 \\
        & CIFAR-10(10\% Noise) & 0.964$\pm$  0.005 \\
        & CIFAR-10(20\% Noise) & 0.881$\pm$  0.013 \\
        & CIFAR-10(30\% Noise) & 0.774$\pm$  0.020 \\
        & CIFAR-10(40\% Noise) & 0.667$\pm$  0.026 \\ \cline{2-3} 
        \multirow{7}{*}{CINIC-10} & STL-10 & 0.940$\pm$  0.047 \\
        & CIFAR-10 & 0.942$\pm$  0.055 \\
        & CINIC-10 & 0.999$\pm$  0.000 \\
        & CINIC-10(10\% Noise) & 0.967$\pm$  0.003 \\
        & CINIC-10(20\% Noise) & 0.884$\pm$  0.011 \\
        & CINIC-10(30\% Noise) & 0.776$\pm$  0.017 \\
        & CINIC-10(40\% Noise) & 0.667$\pm$  0.022 \\ \cline{2-3} 
        \multirow{7}{*}{STL-10} & CIFAR-10 & 0.894$\pm$  0.094 \\
        & CINIC-10 & 0.939$\pm$  0.061 \\
        & STL-10 & 0.997$\pm$  0.000 \\
        & STL-10(10\% Noise) & 0.965$\pm$  0.009 \\
        & STL-10(20\% Noise) & 0.880$\pm$  0.021 \\
        & STL-10(30\% Noise) & 0.770$\pm$  0.030 \\
        & STL-10(40\% Noise) & 0.660$\pm$  0.035 \\ \hline
    \end{tabular}
    % }
    \label{tab:cifar-cinic-stl-C}
\end{table}

\subsubsection{\textbf{Credibility on CIFAR-10, CINIC-10 \& STL-10}}
Intra-dataset CREDSs (Table~\ref{tab:cifar-cinic-stl-C}) are high: $0.998$, $0.999$, and $0.997$ for CIFAR-10, CINIC-10, and STL-10 respectively. CIFAR-10/CINIC-10 cross-evaluation yields $\sim$0.94, as CINIC-10 contains CIFAR-10 images, making prior probabilities $p(X_{k}^t/Y=j)$ similar. However, CIFAR-10/STL-10 cross-evaluation drops to $\approx$ 0.89, indicating dissimilar feature distributions despite visual similarity.

\subsubsection{\textbf{Impact of Noise on Credibility}}
Addition of Gaussian noise $N(0,1)$ deviates the posterior probability $p(X^v_k|Y=j)$ from the prior $p(X^t_k|Y=j)$, decreasing credibility $C_{k,j}$ (Tables~\ref{tab:MNIST-SVHN-C} and~\ref{tab:cifar-cinic-stl-C}). MNIST exhibits larger credibility drops than SVHN due to simpler learned features being more susceptible to noise.

\begin{table}[t]
    \caption{{$C^\mu_{max}$}, {$C^\sigma_{max}$}, $AUC_{\mu_v }^5$ and $AUC_{\sigma_v }^5$ on MNIST and SVHN (with mean and std. on multiple runs). {$C^\mu_{max}$}, {$C^\sigma_{max}$}, $AUC_{\mu_v }^5$ and $AUC_{\sigma_v }^5$ are estimated for all the permutations of SVHN and MNIST. Noise was varied from 0\% to 50 \% to observe how noise effects {$C^\mu_{max}$}, {$C^\sigma_{max}$}, $AUC_{\mu_v }^5$ and $AUC_{\sigma_v }^5$. }
    \centering
    % \resizebox{\columnwidth}{!}{%
        \begin{tabular}{|l|l|l|l|l|l|}
            \hline
            Train Dataset & Test Dataset & $C^\mu_{max}$ & $C^\sigma_{max}$ & $AUC_{\mu_v }^5$ & $AUC_{\sigma_v }^5$\\ \hline
            \multirow{6}{*}{MNIST} & SVHN & 0.632$\pm$  0.091 & 0.536$\pm$  0.093 & 1.715$\pm$  0.202 & 1.678$\pm$  0.274 \\
            & MNIST & 0.756$\pm$  0.103 & 0.869$\pm$  0.071 & 1.885$\pm$  0.339 & 2.314$\pm$  0.259 \\
            & MNIST(10\% Noise) & 0.762$\pm$  0.091 & 0.805$\pm$  0.074 & 1.953$\pm$  0.329 & 2.199$\pm$  0.261 \\
            & MNIST(20\% Noise) & 0.749$\pm$  0.096 & 0.716$\pm$  0.069 & 2.031$\pm$  0.332 & 2.041$\pm$  0.248 \\
            & MNIST(30\% Noise) & 0.747$\pm$  0.094 & 0.637$\pm$  0.068 & 2.119$\pm$  0.337 & 1.890$\pm$  0.239 \\
            & MNIST(40\% Noise) & 0.735$\pm$  0.096 & 0.557$\pm$  0.064 & 2.199$\pm$  0.350 & 1.723$\pm$  0.226 \\ \cline{2-6} 
            \multirow{6}{*}{SVHN} & MNIST & 0.687$\pm$  0.068 & 0.565$\pm$  0.039 & 2.067$\pm$  0.244 & 1.963$\pm$  0.115 \\
            & SVHN & 0.987$\pm$  0.003 & 0.996$\pm$  0.001 & 2.705$\pm$  0.083 & 2.750$\pm$  0.023 \\
            & SVHN(10\% Noise) & 0.986$\pm$  0.003 & 0.948$\pm$  0.007 & 2.728$\pm$  0.081 & 2.671$\pm$  0.030 \\
            & SVHN(20\% Noise) & 0.980$\pm$  0.003 & 0.851$\pm$  0.011 & 2.801$\pm$  0.078 & 2.498$\pm$  0.036 \\
            & SVHN(30\% Noise) & 0.968$\pm$  0.003 & 0.743$\pm$  0.012 & 2.904$\pm$  0.072 & 2.283$\pm$  0.039 \\
            & SVHN(40\% Noise) & 0.947$\pm$  0.003 & 0.645$\pm$  0.012 & 3.026$\pm$  0.067 & 2.062$\pm$  0.039 \\ \hline
        \end{tabular}
        % }
    \label{tab:MNIST-SVHN-area}
\end{table}

\subsubsection{\textbf{Credibility Curves}}
\label{sec:curves-results}
Fig.~\ref{fig:curves} shows credibility curves for $\mu_v$ and $\sigma_v$ variations. MNIST-trained models show lower $C_{j,max}^\mu$ and $C_{j,max}^\sigma$ than SVHN-trained models, indicating higher susceptibility to distributional changes. CIFAR-10 and CINIC-10 exhibit similar credibility patterns, reflecting their dataset similarity.

\subsubsection{\textbf{Maximum Credibility Potential of a Model}}
Tables~\ref{tab:MNIST-SVHN-area} and \ref{tab:cifar-cinic-stl-area} demonstrate models can achieve maximum potential credibilities ($C_{max}^\mu$, $C_{max}^\sigma$) exceeding their baseline $C$ on OOD data (e.g., MNIST $\rightarrow$ SVHN baseline $C=0.4581$, but $C_{max}^\mu=0.6323$). Adding Gaussian noise minimally impacts $C_{max}^\mu$ but significantly degrades $C_{max}^\sigma$, showing that noise limits model robustness primarily through variance shifts rather than mean shifts.

\subsubsection{\textbf{Model Generalization on Varying OOD Dataset}}
We compute $AUC^5_{\mu_v}$ and $AUC^5_{\sigma_v}$ (Tables~\ref{tab:MNIST-SVHN-area}, \ref{tab:cifar-cinic-stl-area}) to quantify generalization under parameter variations. As expected, AUC values are higher for structurally similar test sets. With increasing noise, $AUC^5_{\mu_v}$ rises while $AUC^5_{\sigma_v}$ falls, suggesting model credibility remains more consistent under mean shifts than under deviation shifts in noisy environments.

\subsubsection{\textbf{Overall Model Generalization Estimates}}
\begin{figure}[t]
    \centering
    \includegraphics[width=5in]{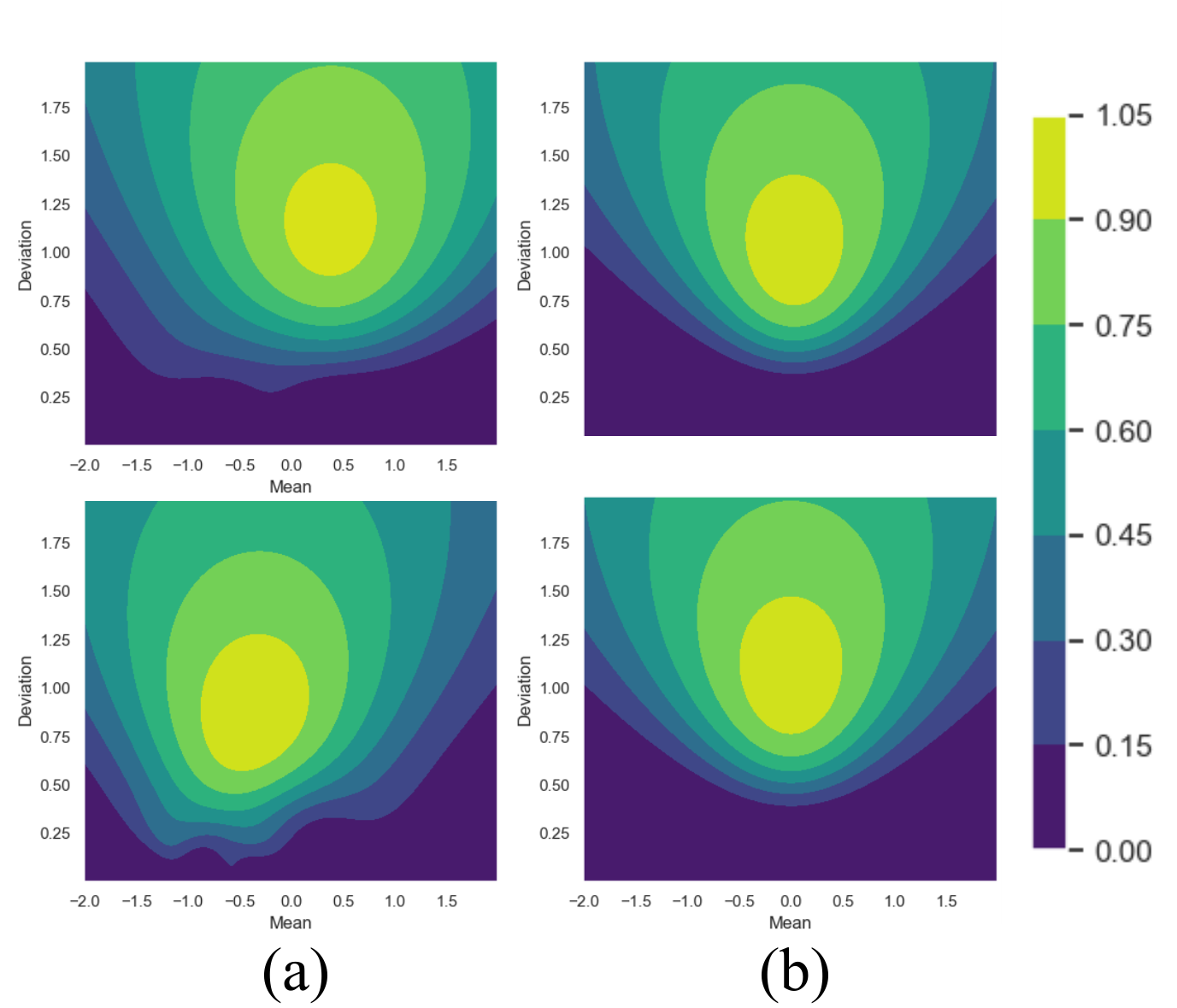}
    \caption{Credibility heatmaps for model trained on (a) MNIST and (b) SVHN for class labels 0 (top) and 1 (bottom).}
    \label{fig:SVHN-MNIST-0-1}
\end{figure}

Credibility heatmaps (Fig.~\ref{fig:SVHN-MNIST-0-1}) reveal class-specific patterns in $C_j$ for simultaneous $\mu_v$ and $\sigma_v$ variations. $VUS^{5,5}$ results (Table~\ref{tab:VUS}) show SVHN models generalize better than MNIST, while CINIC-10 slightly outperforms CIFAR-10 and STL-10 due to ImageNet augmentation and larger sample size.

\begin{table}[t]
    \caption{{$C^\mu_{max}$}, {$C^\sigma_{max}$}, $AUC_{\mu_v }^5$ and $AUC_{\sigma_v }^5$ for various experiments on CIFAR-10, STL-10 and CINIC-10 datasets. Results are reported in the form of mean and deviation across multiple runs. {$C^\mu_{max}$}, {$C^\sigma_{max}$}, $AUC_{\mu_v }^5$ and $AUC_{\sigma_v }^5$ are estimated for all the permutations of CIFAR-10, STL-10 and CINIC-10 dataset. Additionally, we increase noise level in each dataset from 0\% to 50 \% to observe how noise effects {$C^\mu_{max}$}, {$C^\sigma_{max}$}, $AUC_{\mu_v }^5$ and $AUC_{\sigma_v }^5$.}
    \centering
    % \resizebox{\columnwidth}{!}{%
        \begin{tabular}{|l|l|l|l|l|l|}
            \hline
            Train Dataset & Test Dataset & $C^\mu_{max}$ & $C^\sigma_{max}$ & $AUC_{\mu_v }^5$ & $AUC_{\sigma_v }^5$\\ \hline
            \multirow{7}{*}{CIFAR 10} & CINIC-10 & 0.963$\pm$  0.023 & 0.956$\pm$  0.046 & 2.305$\pm$  0.110 & 2.643$\pm$  0.096 \\
            & STL-10 & 0.953$\pm$  0.022 & 0.922$\pm$  0.067 & 2.250$\pm$  0.140 & 2.590$\pm$  0.131 \\
            & CIFAR-10 & 0.976$\pm$  0.019 & 0.995$\pm$  0.002 & 2.387$\pm$  0.175 & 2.705$\pm$  0.080 \\
            & CIFAR-10(10\% Noise) & 0.976$\pm$  0.019 & 0.962$\pm$  0.007 & 2.412$\pm$  0.172 & 2.654$\pm$  0.080 \\
            & CIFAR-10(20\% Noise) & 0.976$\pm$  0.018 & 0.885$\pm$  0.012 & 2.483$\pm$  0.166 & 2.526$\pm$  0.085 \\
            & CIFAR-10(30\% Noise) & 0.973$\pm$  0.017 & 0.791$\pm$  0.017 & 2.585$\pm$  0.161 & 2.354$\pm$  0.088 \\
            & CIFAR-10(40\% Noise) & 0.964$\pm$  0.015 & 0.700$\pm$  0.021 & 2.707$\pm$  0.156 & 2.169$\pm$  0.091 \\ \cline{2-6} 
            \multirow{7}{*}{CINIC-10} & STL-10 & 0.971$\pm$  0.011 & 0.949$\pm$  0.050 & 2.365$\pm$  0.161 & 2.643$\pm$  0.101 \\
            & CIFAR-10 & 0.971$\pm$  0.017 & 0.954$\pm$  0.046 & 2.433$\pm$  0.178 & 2.650$\pm$  0.069 \\
            & CINIC-10 & 0.984$\pm$  0.011 & 0.997$\pm$  0.002 & 2.423$\pm$  0.123 & 2.718$\pm$  0.061 \\
            & CINIC-10(10\% Noise) & 0.984$\pm$  0.011 & 0.965$\pm$  0.003 & 2.449$\pm$  0.121 & 2.668$\pm$  0.064 \\
            & CINIC-10(20\% Noise) & 0.983$\pm$  0.010 & 0.887$\pm$  0.010 & 2.522$\pm$  0.120 & 2.537$\pm$  0.070 \\
            & CINIC-10(30\% Noise) & 0.978$\pm$  0.008 & 0.791$\pm$  0.016 & 2.629$\pm$  0.116 & 2.360$\pm$  0.073 \\
            & CINIC-10(40\% Noise) & 0.967$\pm$  0.007 & 0.697$\pm$  0.019 & 2.754$\pm$  0.114 & 2.169$\pm$  0.075 \\ \cline{2-6} 
            \multirow{7}{*}{STL-10} & CIFAR-10 & 0.950$\pm$  0.035 & 0.915$\pm$  0.075 & 2.409$\pm$  0.170 & 2.558$\pm$  0.099 \\
            & CINIC-10 & 0.958$\pm$  0.030 & 0.951$\pm$  0.048 & 2.378$\pm$  0.128 & 2.619$\pm$  0.074 \\
            & STL-10 & 0.966$\pm$  0.029 & 0.993$\pm$  0.002 & 2.356$\pm$  0.151 & 2.686$\pm$  0.080 \\
            & STL-10(10\% Noise) & 0.967$\pm$  0.028 & 0.962$\pm$  0.009 & 2.382$\pm$  0.149 & 2.638$\pm$  0.087 \\
            & STL-10(20\% Noise) & 0.967$\pm$  0.027 & 0.884$\pm$  0.018 & 2.454$\pm$  0.148 & 2.509$\pm$  0.096 \\
            & STL-10(30\% Noise) & 0.964$\pm$  0.025 & 0.787$\pm$  0.026 & 2.559$\pm$  0.144 & 2.333$\pm$  0.104 \\
            & STL-10(40\% Noise) & 0.954$\pm$  0.023 & 0.693$\pm$  0.029 & 2.685$\pm$  0.142 & 2.144$\pm$  0.105 \\ \hline
        \end{tabular}
        % }
    \label{tab:cifar-cinic-stl-area}
\end{table}

\begin{table}[t]
    \caption{$VUS^{5,5}$ calculated for models trained on MNIST, SVHN, CIFAR-10, STL-10 and CINIC-10, with mean and deviation across multiple runs.}
    \centering
    % \resizebox{.45\columnwidth}{!}{%
    \begin{tabular}{|l|l|}
        \hline
        Train Dataset & $VUS^{5,5}$ \\ \hline
        MNIST & 10.99$\pm$  1.62 \\
        SVHN & 13.93$\pm$  0.19 \\
        CIFAR-10 & 13.55$\pm$  0.66 \\
        STL-10 & 13.39$\pm$  0.65 \\
        CINIC-10 & 13.64$\pm$  0.50 \\ \hline
    \end{tabular}
    % }
    \label{tab:VUS}
\end{table}

\section{Conclusion}
This work introduces PLSP, a paradigm shift from post-hoc OOD detection to pre-hoc OOD prediction through the CREDibility Score (CREDS) metric. Unlike existing approaches requiring OOD datasets during inference, CREDS enables anticipatory assessment of model behavior under distributional shifts during development. Experiments across MNIST, SVHN, CIFAR-10, CINIC-10, and STL-10 demonstrate that CREDS provides estimates comparable to post-hoc accuracy metrics while offering evaluation in the absence of OOD data. The introduced credibility curves, heatmaps, and VUS metrics provide comprehensive tools for analyzing model vulnerability, revealing differential sensitivity to mean versus variance shifts. The strong correlation between CREDS and accuracy validates its utility as a pre-deployment robustness indicator, enabling proactive adjustments to prevent catastrophic failures in production environments.

% \bibliographystyle{unsrt}  
% \bibliography{egbib}  %%% Remove comment to use the external .bib file (using bibtex).
% %%% and comment out the ``thebibliography'' section.

%%% Comment out this section when you \bibliography{egbib} is enabled. Following section contains the output from .bbl file.

\begin{thebibliography}{10}

\bibitem{yang2021generalized}
Jingkang Yang, Kaiyang Zhou, Yixuan Li, and Ziwei Liu.
\newblock Generalized out-of-distribution detection: A survey.
\newblock {\em arXiv preprint arXiv:2110.11334}, 2021.

\bibitem{roy2021does}
Abhijit~Guha Roy, Jie Ren, Shekoofeh Azizi, Aaron Loh, Vivek Natarajan, Basil Mustafa, Nick Pawlowski, Jan Freyberg, Yuan Liu, Zach Beaver, et~al.
\newblock Does your dermatology classifier know what it doesn't know? detecting the long-tail of unseen conditions.
\newblock {\em arXiv preprint arXiv:2104.03829}, 2021.

\bibitem{gulrajani2020search}
Ishaan Gulrajani and David Lopez-Paz.
\newblock In search of lost domain generalization.
\newblock In {\em International Conference on Learning Representations}, 2020.

\bibitem{fort2021exploring}
Stanislav Fort, Jie Ren, and Balaji Lakshminarayanan.
\newblock Exploring the limits of out-of-distribution detection.
\newblock {\em arXiv preprint arXiv:2106.03004}, 2021.

\bibitem{krizhevsky2012imagenet}
Alex Krizhevsky, Ilya Sutskever, and Geoffrey~E Hinton.
\newblock Imagenet classification with deep convolutional neural networks.
\newblock {\em Advances in neural information processing systems}, 25:1097--1105, 2012.

\bibitem{he2015delving}
Kaiming He, Xiangyu Zhang, Shaoqing Ren, and Jian Sun.
\newblock Delving deep into rectifiers: Surpassing human-level performance on imagenet classification.
\newblock In {\em Proceedings of the IEEE international conference on computer vision}, pages 1026--1034, 2015.

\bibitem{drummond2006open}
Nick Drummond and Rob Shearer.
\newblock The open world assumption.
\newblock In {\em eSI Workshop: The Closed World of Databases meets the Open World of the Semantic Web}, volume~15, 2006.

\bibitem{liu1995unbiased}
Yong Liu.
\newblock Unbiased estimate of generalization error and model selection in neural network.
\newblock {\em Neural Networks}, 8(2):215--219, 1995.

\bibitem{winkens2020contrastive}
Jim Winkens, Rudy Bunel, Abhijit~Guha Roy, Robert Stanforth, Vivek Natarajan, Joseph~R Ledsam, Patricia MacWilliams, Pushmeet Kohli, Alan Karthikesalingam, Simon Kohl, et~al.
\newblock Contrastive training for improved out-of-distribution detection.
\newblock {\em arXiv preprint arXiv:2007.05566}, 2020.

\bibitem{milbich2021characterizing}
Timo Milbich, Karsten Roth, Samarth Sinha, Ludwig Schmidt, Marzyeh Ghassemi, and Bj{\"o}rn Ommer.
\newblock Characterizing generalization under out-of-distribution shifts in deep metric learning.
\newblock In {\em Advances in Neural Information Processing Systems}, 2021.

\bibitem{thanh2020toward}
Hoang Thanh-Tung and Truyen Tran.
\newblock Toward a generalization metric for deep generative models.
\newblock In {\em ''I Can't Believe It's Not Better!''NeurIPS 2020 workshop}, 2020.

\bibitem{frogner2021incorporating}
Charlie Frogner, Sebastian Claici, Edward Chien, and Justin Solomon.
\newblock Incorporating unlabeled data into distributionally robust learning.
\newblock {\em Journal of Machine Learning Research}, 22(56):1--46, 2021.

\bibitem{shen2020stable}
Zheyan Shen, Peng Cui, Tong Zhang, and Kun Kunag.
\newblock Stable learning via sample reweighting.
\newblock In {\em Proceedings of the AAAI Conference on Artificial Intelligence}, volume~34, pages 5692--5699, 2020.

\bibitem{liu2021heterogeneous}
Jiashuo Liu, Zheyuan Hu, Peng Cui, Bo~Li, and Zheyan Shen.
\newblock Heterogeneous risk minimization.
\newblock {\em arXiv preprint arXiv:2105.03818}, 2021.

\bibitem{zhou2021domain}
Kaiyang Zhou, Ziwei Liu, Yu~Qiao, Tao Xiang, and Chen~Change Loy.
\newblock Domain generalization: A survey.
\newblock {\em arXiv preprint arXiv:2103.02503}, 2021.

\bibitem{ye2021out}
Haotian Ye, Chuanlong Xie, Yue Liu, and Zhenguo Li.
\newblock Out-of-distribution generalization analysis via influence function.
\newblock {\em arXiv preprint arXiv:2101.08521}, 2021.

\bibitem{ye2021ood}
Nanyang Ye, Kaican Li, Lanqing Hong, Haoyue Bai, Yiting Chen, Fengwei Zhou, and Zhenguo Li.
\newblock Ood-bench: Benchmarking and understanding out-of-distribution generalization datasets and algorithms.
\newblock {\em arXiv preprint arXiv:2106.03721}, 2021.

\bibitem{neyshabur2017exploring}
Behnam Neyshabur, Srinadh Bhojanapalli, David Mcallester, and Nati Srebro.
\newblock Exploring generalization in deep learning.
\newblock {\em Advances in Neural Information Processing Systems}, 30:5947--5956, 2017.

\bibitem{arjovsky2020invariant}
Martin Arjovsky, L{\'e}on Bottou, Ishaan Gulrajani, and David Lopez-Paz.
\newblock Invariant risk minimization.
\newblock {\em stat}, 1050:27, 2020.

\bibitem{krueger2021out}
David Krueger, Ethan Caballero, Joern-Henrik Jacobsen, Amy Zhang, Jonathan Binas, Dinghuai Zhang, Remi Le~Priol, and Aaron Courville.
\newblock Out-of-distribution generalization via risk extrapolation (rex).
\newblock In {\em International Conference on Machine Learning}, pages 5815--5826. PMLR, 2021.

\bibitem{xu2020adversarial}
Minghao Xu, Jian Zhang, Bingbing Ni, Teng Li, Chengjie Wang, Qi~Tian, and Wenjun Zhang.
\newblock Adversarial domain adaptation with domain mixup.
\newblock In {\em Proceedings of the AAAI Conference on Artificial Intelligence}, volume~34, pages 6502--6509, 2020.

\bibitem{yan2020improve}
Shen Yan, Huan Song, Nanxiang Li, Lincan Zou, and Liu Ren.
\newblock Improve unsupervised domain adaptation with mixup training.
\newblock {\em arXiv preprint arXiv:2001.00677}, 2020.

\bibitem{Kullback59}
Solomon Kullback.
\newblock {\em Information Theory and Statistics}.
\newblock Wiley, New York, 1959.

\bibitem{zhang2021properties}
Yufeng Zhang, Wanwei Liu, Zhenbang Chen, Kenli Li, and Ji~Wang.
\newblock On the properties of kullback-leibler divergence between gaussians.
\newblock {\em arXiv preprint arXiv:2102.05485}, 2021.

\bibitem{MNIST}
Li~Deng.
\newblock The mnist database of handwritten digit images for machine learning research [best of the web].
\newblock {\em IEEE Signal Processing Magazine}, 29(6):141--142, 2012.

\bibitem{SVHN}
Yuval Netzer, Tao Wang, Adam Coates, Alessandro Bissacco, Bo~Wu, and Andrew~Y. Ng.
\newblock Reading digits in natural images with unsupervised feature learning.
\newblock In {\em NIPS Workshop on Deep Learning and Unsupervised Feature Learning 2011}, 2011.

\bibitem{cifar}
Benjamin Recht, Rebecca Roelofs, Ludwig Schmidt, and Vaishaal Shankar.
\newblock Do cifar-10 classifiers generalize to cifar-10?
\newblock {\em arXiv preprint arXiv:1806.00451}, 2018.

\bibitem{CINIC}
Luke~N Darlow, Elliot~J Crowley, Antreas Antoniou, and Amos~J Storkey.
\newblock Cinic-10 is not imagenet or cifar-10.
\newblock {\em arXiv e-prints}, pages arXiv--1810, 2018.

\bibitem{STL}
Adam Coates, Honglak Lee, and Andrew~Y. Ng.
\newblock An analysis of single layer networks in unsupervised feature learning.
\newblock 2011.

\bibitem{RF}
Leo Breiman.
\newblock Random forests.
\newblock {\em Machine learning}, 45(1):5--32, 2001.

\end{thebibliography}

\end{document}